\documentclass[10pt,twocolumn,letterpaper]{article}
\usepackage[pagenumbers]{cvpr}
\usepackage{listings}
\definecolor{cvprblue}{rgb}{0.21,0.49,0.74}
\usepackage[pagebackref,breaklinks,colorlinks,allcolors=cvprblue]{hyperref}
\def\confName{CVPR}
\def\confYear{2027}
\title{EverybodyDance: Music-Driven Video Generation for Diverse Characters}

\author{
Yan Zhang\thanks{Equal contribution.} \quad Han Zou\footnotemark[1] \quad
Cong Xie \quad Lincong Feng \quad Ruiqi Yu \quad Zhenpeng Zhan\\
Baidu Inc.
}

\begin{document}
\maketitle
\begin{abstract}
Generating a dance video from music and a single reference image requires natural, music-responsive dancing and coherent appearance over time. Extending this beyond realistic humans is difficult because paired music--dance videos of stylized characters are scarce. We present \textbf{EverybodyDance}, a modular framework that connects music-conditioned keypoint-sequence generation with a pretrained pose-conditioned video generation model through an explicit 2D keypoint interface. Because 2D keypoints decouple body pose from visual texture and rendering style, the first stage can learn music--dance relationships solely from in-the-wild real-human videos, while the video generation model renders diverse characters and accommodates moderate differences in body proportions. To generate reliable keypoint sequences, we encode each coordinate along a spatial-bin axis and stack the encodings over time, turning keypoint-sequence generation into music-conditioned trajectory-image synthesis, analogous to text-to-image generation. A trajectory-image VAE trained from scratch classifies each coordinate over spatial bins plus a null class, recovering both locations and missing landmarks. A flow Transformer generates its latents, conditioned on music tokens that share temporal indices with the trajectory and on the separately encoded reference pose. On references spanning realistic humans, anime characters, and anthropomorphic animals, EverybodyDance achieves higher video-level scores and user preference than X-Dancer, MusicInfuser, and Wan-Dancer, including X-Dancer paired with the same pose-conditioned video generation model. See \url{https://hot-dance.github.io/} for our generated dance videos of diverse characters.
\end{abstract}

\section{Introduction}

Generating a dance video from arbitrary music and a single reference image is
an appealing yet challenging problem in controllable video generation. The
video must contain natural and diverse dance dynamics, respond to both local
rhythms and broader musical changes, and remain visually convincing over time.
Practical systems should moreover extend beyond realistic humans to anime
figures, anthropomorphic animals, and other stylized characters. This setting
is particularly challenging because large collections of high-quality paired
music--dance videos overwhelmingly depict real humans, whereas comparable data
for stylized characters are scarce.

Pose-conditioned video generation models provide a practical second stage for
music-driven video synthesis \citep{Animate-anyone,Wan-Animate,yan2026scail}.
Given a reference image and a 2D keypoint sequence, a compatible model can
generate videos of realistic humans and stylized characters. The remaining challenge is
to generate suitable 2D keypoint sequences from music. These sequences serve as
structural controls, while the video generation model handles character
appearance synthesis. This separation allows music--dance relationships to be
learned from real-human videos without training the keypoint-sequence generator
on paired dance videos from each target character domain.
Recent music-driven systems such as X-Dancer \citep{chen2025xdancer},
Wan-Dancer \citep{huang2026wandancer}, and MusicInfuser \citep{hong2026musicinfuser}
do not use this separation: X-Dancer passes learned motion features to its own
video generation model, whereas Wan-Dancer and MusicInfuser learn music
alignment within the video generation model.
However, pronounced jitter, implausible configurations, or substantial geometric
drift in the generated keypoint sequences can propagate into unstable limbs,
body deformation, and temporal flicker during video synthesis.
Reliable keypoint-sequence generation and modular integration must therefore be addressed
together.

We present \textbf{EverybodyDance}, a modular framework for music-driven video
generation across diverse characters (Figure~\ref{fig:method_overview}).
Its two stages communicate through explicit 2D keypoint sequences.
The first stage learns from large-scale monocular
real-human dance videos and generates keypoint coordinates and validity
information from music. The second stage uses a pretrained pose-conditioned
video generation model to synthesize the video from the generated keypoint
sequence and a reference image.

Generating such keypoint sequences from in-the-wild data remains difficult. The training
sequences vary substantially in dance style, body proportions, framing,
occlusion, and keypoint visibility. Directly generating raw 2D coordinate
sequences produces implausible body poses, unnatural limb geometry, and
structural distortion over time. Sequences generated in the latent space of a matched raw-coordinate
autoencoder exhibit abrupt changes in character scale and floating feet
(Section~\ref{sec:control_ablation}). To address these failures,
we reframe keypoint-sequence generation as music-token-conditioned multi-channel image
synthesis: keypoint locations are encoded along a spatial-bin axis and stacked
over time to form trajectory images. This yields a token-conditioned image
synthesis problem analogous to text-to-image generation, with music tokens
conditioning the synthesis of trajectory images.
A domain-specific autoencoder, trained from scratch with
categorical coordinate and visibility targets, compresses the trajectory images
for flow-based generation. Shared temporal indices align each music token with
its trajectory time step, and the reference pose is encoded separately so that
the generated sequence starts from the pose in the reference image.

We evaluate EverybodyDance using in-the-wild music and reference images
spanning realistic humans, anime characters, and anthropomorphic animals. Video-level assessment
and a 47-rater user study favor our system over X-Dancer, MusicInfuser, and
Wan-Dancer in evaluated video quality and overall preference.
The advantage persists when X-Dancer's first-stage output is used to condition
video generation with the same SCAIL \citep{yan2026scail} backbone and inference settings as
EverybodyDance, supporting the value of our first-stage design beyond the
choice of video generation backbone.

In summary, our contributions are threefold.
(1)~\textbf{A modular framework for music-driven character video generation.}
EverybodyDance connects music-conditioned keypoint-sequence generation
with pretrained pose-conditioned video generation through an explicit 2D
keypoint interface. Because 2D keypoints decouple motion kinematics from visual texture and
rendering style, the first stage can learn music--dance relationships solely
from in-the-wild real-human videos, while the video generation model renders diverse
characters and accommodates moderate differences in body proportions. The same generator thus also serves anime characters
and anthropomorphic animals, without paired music--dance data from those
domains.
(2)~\textbf{Keypoint-sequence generation as trajectory-image synthesis.}
We encode each keypoint coordinate along a spatial-bin axis and stack the
encodings over time, turning music-to-dance generation into
music-token-conditioned image synthesis that can adopt latent image
generation models. With the same DiT backbone and training objective, it
outperforms generating raw coordinates with or without an autoencoder
(Table~\ref{tab:control_ablation}).
(3)~\textbf{A trajectory-image codec and music alignment.}
Our VAE classifies each coordinate over spatial bins plus a null class,
recovering both keypoint locations and missing landmarks
(Table~\ref{tab:codec_ablation}). Music tokens share rotary temporal
indices with the trajectory latents they accompany, which improves beat
alignment and movement coherence over text-to-image-style indexing
(Table~\ref{tab:control_ablation}).

\section{Related Work}
\label{sec:related_work}

\paragraph{Music-driven character video generation.}
X-Dancer \citep{chen2025xdancer} generates compositional 2D body, head, and
hand pose tokens with a music-conditioned autoregressive Transformer; its
selected pipeline decodes them into learned multiscale guidance for a video
diffusion UNet rather than explicit skeletons. Wan-Dancer
\citep{huang2026wandancer} targets minute-scale generation through global
keyframe planning and local temporal refinement, with full-track musical
context and time-mapped rotary embeddings. MusicInfuser
\citep{hong2026musicinfuser} fine-tunes layers of a pretrained text-to-video
diffusion model, selected by a layer-wise adaptability criterion, to learn
music alignment without motion data or an intermediate keypoint generator.
EverybodyDance instead generates explicit 2D keypoint sequences as
music-conditioned trajectory images and passes them to a pretrained
pose-conditioned video generation model.

\paragraph{Pose-conditioned video generation.}
Animate Anyone \citep{Animate-anyone} and MagicAnimate
\citep{xu2024magicanimate} animate a reference image with diffusion models
following supplied motion controls. Wan-Animate \citep{Wan-Animate} unifies
character animation and replacement with skeleton-based body control and
facial expression features, and SCAIL \citep{yan2026scail} introduces a
3D-consistent pose representation with full-context pose injection. These
systems animate a given driving signal but do not decide which dance to
generate from music. EverybodyDance is complementary: its first stage,
trained on real-human dance data, generates this signal from music for such
a backbone.

\paragraph{Music-conditioned dance motion.}
Music-to-dance methods generate motion rather than character videos.
FACT \citep{FACT2} uses a cross-modal Transformer, Bailando \citep{Bailando2}
combines a quantized motion memory with an actor--critic Transformer and a
beat-alignment reward, EDGE \citep{Edge2} applies Transformer-based diffusion
to Jukebox features, and Lodge \citep{Lodge2} links global and local diffusion
models through dance primitives. These methods primarily target 3D dance
sequences. Such 3D data require motion capture or multi-view reconstruction
and are far scarcer than monocular dance videos; AIST++ \citep{FACT2}, for
example, contains 5.2 hours of 3D dance motion. EverybodyDance instead learns
from monocular dance videos to
generate 2D keypoint sequences that condition video synthesis.

\paragraph{Coordinate representations and latent image synthesis.}
Pose estimation represents keypoint locations with continuous coordinates
\citep{Deeppose2,Human_pose_regression2}, 2D heatmaps
\citep{Stacked,Deep-high-resolution}, or per-axis classification over spatial
bins, as in SimCC \citep{li2022simcc}. EverybodyDance also uses spatial-bin
categories, but to generate entire keypoint sequences from music rather than
localize keypoints in an image: it stacks the encodings along time into
trajectory images, with zero inputs and a decoder null class for missing axes.
This connects keypoint-sequence generation to latent image synthesis, including
latent diffusion \citep{rombach2022ldm}, DiT \citep{peebles2023dit},
rectified-flow Transformers \citep{esser2024rectified}, and Lumina-Image 2.0
\citep{Lumina-image-2.0}. We adopt this modeling structure rather than
treating trajectories as RGB images: our codec is trained from scratch with
categorical reconstruction, and music tokens are explicitly aligned to the
generated sequence.

\section{Method}
\label{sec:method}

\subsection{Overview}
\label{sec:overview}

\begin{figure*}[t]
    \centering
    \includegraphics[width=\textwidth]{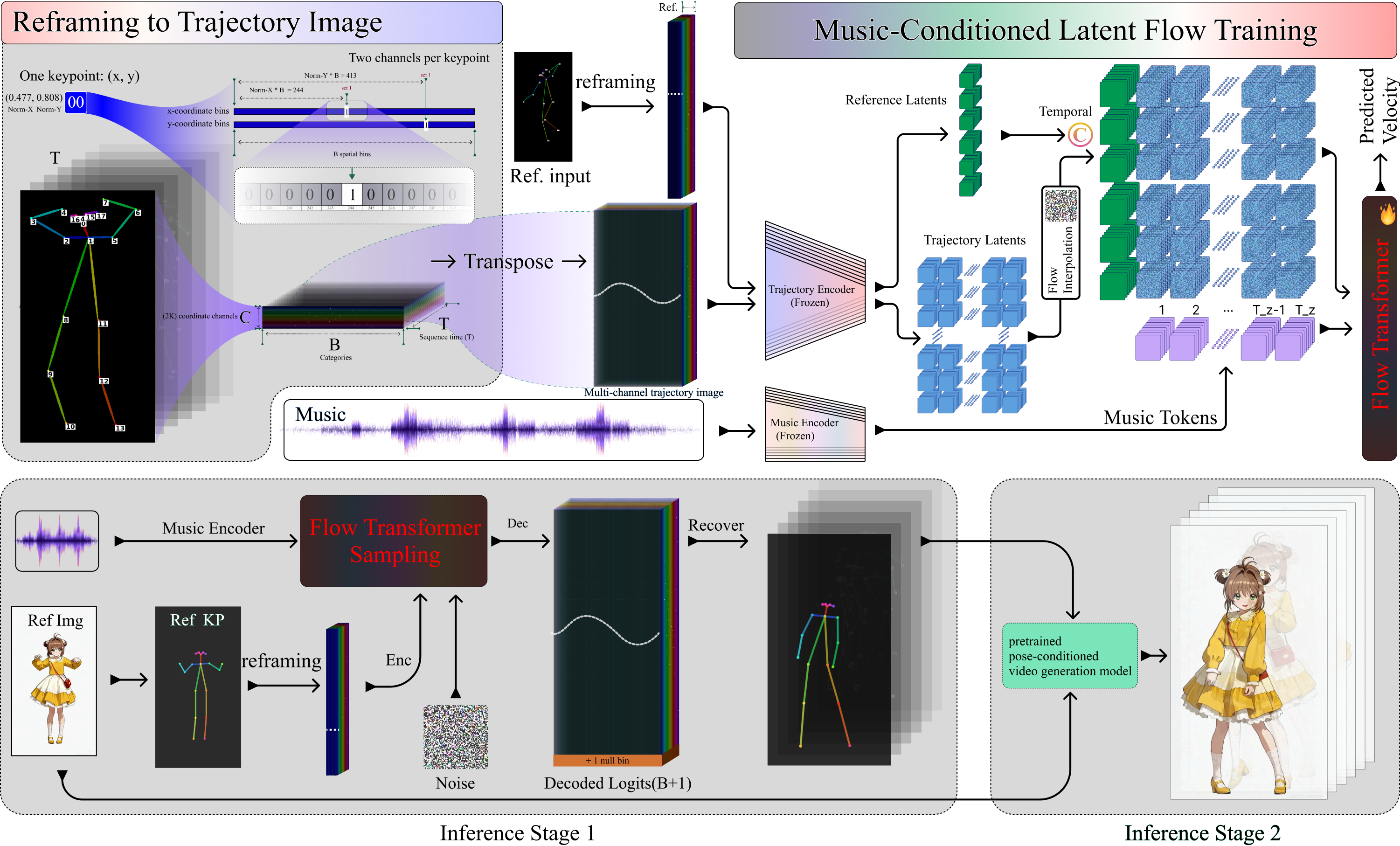}
    \caption{\textbf{Overview of EverybodyDance.}
    \textbf{Upper left:} trajectory-image construction from per-axis one-hot
    spatial bins (zero vectors for missing axes).
    \textbf{Upper right:} flow training, where clean reference latents are
    prepended in time to noisy target latents and each music token is aligned
    with one latent time step; the VAE and music encoder are frozen.
    \textbf{Bottom:} inference, where decoded keypoint sequences and the
    reference image drive a fixed, pretrained pose-conditioned video generation
    model.}
    \label{fig:method_overview}
\end{figure*}

Given a music waveform $m$ and a reference image $I_{\mathrm{ref}}$,
EverybodyDance generates a video of the depicted character dancing to the music.
The bottom panels of Figure~\ref{fig:method_overview} illustrate the overall
two-stage pipeline.
In our pipeline, the second stage uses a 2D keypoint sequence, including
coordinates and validity indicators, as structural control. The first stage
therefore generates this sequence $\hat X=G_{\theta}(m,x_{\mathrm{ref}})$ from music and reference keypoints
$x_{\mathrm{ref}}$, and the second stage synthesizes the video $\hat V=R_{\omega}(I_{\mathrm{ref}},\hat X)$ with the fixed, pretrained pose-conditioned video generation model $R_{\omega}$.

To generate reliable 2D keypoint sequences in the first stage, we represent
each sequence as a multi-channel trajectory image and encode the accompanying
music into a sequence of tokens.
A dedicated trajectory-image VAE compresses these images into latents, and a
flow Transformer generates the latents conditioned on music. The VAE decoder
predicts spatial-coordinate and null logits from the generated latents. These
logits are used to recover keypoint coordinates and validity indicators for
video synthesis.

\subsection{Reframing Keypoint Generation as Multi-Channel Image Synthesis}
\label{sec:reframing}

The first stage must generate an entire 2D keypoint sequence from music.
A central design choice is how to represent this sequence for conditional
generation. The most direct formulation models sequences of raw 2D coordinate
vectors, without converting them into trajectory images. In our in-the-wild
experiments (Section~\ref{sec:control_ablation}), however, this Raw Coordinates
variant produces implausible body poses, unnatural limb geometry, and
structural distortion over time, and the Raw Coordinates + VAE variant exhibits
abrupt changes in character scale and floating feet.
These observations motivate an alternative formulation that connects
keypoint-sequence generation to the mature tools of conditional image synthesis.

A straightforward alternative is to formulate each pose as a skeleton image.
However, each image captures only a single frame: representing an entire keypoint
sequence therefore requires a sequence of skeleton images, turning the task into
video generation rather than image generation, making this approach computationally
expensive, especially at high spatial resolutions and for long sequences.
We take a different route: \emph{rather than drawing the body at each time
step, we construct an image that records how each coordinate changes over
the entire sequence}. Each coordinate is encoded along a spatial-bin axis,
and successive frames occupy successive columns along the other axis.
The target is thus a multi-channel position--time image rather than a
skeleton video, with one channel for each keypoint coordinate.

\paragraph{Reframing keypoint sequences as trajectory images.}
Let $X$ contain $K$ keypoints over $T$ frames, with normalized coordinates
$(x_{k,t},y_{k,t})\in[0,1]^2$ for keypoint $k$ at frame $t$.
Body joints use image coordinates; hand and facial landmarks use
wrist-relative coordinates with an asinh warp and nose-relative coordinates
with linear normalization, respectively (Appendix~\ref*{app:finegrained}).
The annotation indicator $v^a_{k,t}\in\{0,1\}$ specifies whether coordinate
axis $a\in\{x,y\}$ is valid. We use $B=512$ spatial bins, indexed
$0,\ldots,B-1$, and $C=2K$ coordinate channels.

Consider a valid horizontal coordinate $x_{k,t}$. We assign it to a spatial bin:
\begin{equation}
    b^x_{k,t}=
    \min\!\left(B-1,\lfloor Bx_{k,t}\rfloor\right),
    \qquad v^x_{k,t}=1.
    \label{eq:bin_assignment}
\end{equation}
Its length-$B$ encoding is
\begin{equation}
    \mathbf h^x_{k,t}[b]=
    \begin{cases}
        1, & v^x_{k,t}=1\ \text{and}\ b=b^x_{k,t},\\
        0, & \text{otherwise},
    \end{cases}
    \quad b=0,\ldots,B-1.
    \label{eq:categorical_vector}
\end{equation}
We apply the same construction to the vertical coordinate. Thus, a valid
axis is represented by a one-hot vector, whereas a missing axis is represented
by an all-zero vector.

For a single frame, we stack the coordinate encodings in the channel order
$(x_1,y_1,\ldots,x_K,y_K)$:
\begin{equation}
    \begin{gathered}
        Q_t =
        \big[
            \mathbf h^x_{1,t},\mathbf h^y_{1,t},
            \ldots,\mathbf h^x_{K,t},\mathbf h^y_{K,t}
        \big]^{\top},\\
        Q_t\in\{0,1\}^{2K\times B}.
    \end{gathered}
    \label{eq:frame_encoding}
\end{equation}
Stacking frames along the temporal axis gives the trajectory image
\begin{equation}
    \begin{gathered}
        H(X)[:,:,t]=Q_t,\\
        H(X)\in\{0,1\}^{C\times B\times T}.
    \end{gathered}
    \label{eq:trajectory_image}
\end{equation}
With channel indices $c_x(k)=2k-1$ and $c_y(k)=2k$,
$H(X)[c_a(k),b,t]=\mathbf h^a_{k,t}[b]$.
Each channel is a $B\times T$ position--time image: rows index spatial bins
and columns index frames. As a coordinate changes over time, its active bin
traces a trajectory across the image; a missing coordinate leaves a zero
column.

The procedure for recovering keypoint sequences from trajectory images is
given in Appendix~\ref*{app:keypoint_recovery}.

\paragraph{Connecting to the text-to-image toolchain.}
This reformulation allows us to leverage the modern toolchain of DiT-style,
token-conditioned image generation
\citep{rombach2022ldm,peebles2023dit,esser2024rectified}.
These models have demonstrated strong capacity to capture complex, highly
varied image distributions, motivating their use for the high-variance dance
distributions encountered in the wild. This formulation is analogous to text-to-image
synthesis: instead of generating an image conditioned on text tokens, we
generate a trajectory image conditioned on music tokens. We can therefore
adapt the modeling framework of token-conditioned latent image generation. This shared formulation also
provides a route to incorporate further advances in conditional image
generation without redefining the underlying trajectory image representation.
We retain this generation structure and introduce three trajectory-specific
adaptations: a trajectory-image VAE with categorical reconstruction
(Section~\ref{sec:codec}),
music--trajectory temporal alignment (Section~\ref{sec:alignment}), and reference
trajectory conditioning (Section~\ref{sec:reference}).

\subsection{Learning a Trajectory-Image VAE}
\label{sec:codec}

We learn a dedicated variational autoencoder (VAE), which serves as our
trajectory codec, to compress trajectory images into a latent space for
subsequent music-conditioned generation. Unlike RGB
images, trajectory images use separate channels for keypoint coordinate axes,
with spatial bins and sequence time forming the two image axes. The codec
therefore reconstructs coordinate locations and missingness rather than color
intensities. We train an encoder $E_{\phi}$ and decoder $D_{\psi}$ from scratch
using the trajectory-image inputs defined in Section~\ref{sec:reframing}.

The encoder parameterizes a diagonal Gaussian posterior over trajectory
latents $Z\in\mathbb{R}^{C_z\times T_z\times B_z}$, retaining a grid over
sequence time and spatial position. In our configuration, both axes are
downsampled by a factor of eight: $T_z=T/8$ and $B_z=B/8=64$.

The decoder maps these latents to coordinate-category logits, as described below.

\paragraph{Supervising coordinate locations and missingness.}
For each coordinate axis at each frame, the decoder predicts a distribution
over 512 spatial bins and one null category. Valid coordinates are supervised
with spatial labels, while missing coordinates are supervised with the null
label. We use a cross-entropy reconstruction loss; the full training objective
and loss definitions are provided in Appendix~\ref*{app:vae_training}.

\subsection{Music--Trajectory Temporal Alignment}
\label{sec:alignment}

\paragraph{Music tokens as conditioning tokens.}
We extract frozen Jukebox features \citep{dhariwal2020jukebox} and resample
them to match the VAE's latent temporal rate of $r/8$ Hz (Appendix~\ref*{app:experimental_details}).
For our $T$-frame segments, the resulting music sequence has $T_z=T/8$ tokens,
$A\in\mathbb R^{T_z\times d_a}$, with one music token aligned to each
trajectory-latent time step.

These music features serve as conditioning tokens, analogous to text
embeddings in text-to-image synthesis. The flow Transformer receives the
main trajectory latents and the separately encoded reference latents
described in Section~\ref{sec:reference}, together with the music tokens.

\paragraph{Shared temporal indices.}
To represent this correspondence in the model, we give aligned music and
trajectory tokens the same temporal coordinate in rotary positional encoding
\citep{su2024roformer}:
\begin{equation}
    \begin{aligned}
        \operatorname{pos}_{\mathrm{main}}(i,j)&=(0,i,j),\\
        \operatorname{pos}_{\mathrm{music}}(j)&=(0,0,j),
    \end{aligned}
    \label{eq:shared_index}
\end{equation}
Here $i$ indexes spatial positions and $j$ indexes latent time. For each $j$,
the music token and all trajectory tokens at that time share the temporal
coordinate $j$, explicitly identifying their correspondence in the positional
encoding.

\subsection{Reference Trajectory Conditioning}
\label{sec:reference}

\paragraph{Encoding the reference keypoints.}
During training, the reference keypoints are the first target frame,
$x_{\mathrm{ref}}=X_1$, which remains in the generation target and participates
in supervision so the generated sequence starts from the reference pose.
At inference, $x_{\mathrm{ref}}$ is detected from the reference image. If
detection fails on a stylized character, a generic template skeleton can serve
as $x_{\mathrm{ref}}$ instead, since the pose-conditioned video model tolerates
moderate misalignment between the driving keypoints and the reference image
(Appendix~\ref*{app:template_reference}, Video~S2).
We repeat $x_{\mathrm{ref}}$ for $T_{\mathrm{ref}}$ frames to form a static
reference trajectory image $X_{\mathrm{ref}}=\operatorname{Repeat}(x_{\mathrm{ref}},T_{\mathrm{ref}})$,
and encode it and the complete $T$-frame target independently using the same
VAE encoder:
\begin{equation}
    Z_{\mathrm{ref}} = E_{\phi}\!\left(H(X_{\mathrm{ref}})\right),
    \qquad
    Z_{\mathrm{main}} = E_{\phi}\!\left(H(X_{1:T})\right).
    \label{eq:reference}
\end{equation}
Here $Z_{\mathrm{ref}}\in\mathbb R^{C_z\times T_z^{\mathrm{ref}}\times B_z}$
remains clean and serves only as conditioning, whereas
flow interpolation and prediction losses apply exclusively to the main sequence
$Z_{\tau}\in\mathbb R^{C_z\times T_z\times B_z}$. At each flow time $\tau$,
the trajectory-latent input to the Transformer concatenates them along the latent sequence-time axis,
$\widetilde Z_{\tau}=\operatorname{concat}_{\mathrm{time}}(Z_{\mathrm{ref}},Z_{\tau})
\in\mathbb R^{C_z\times(T_z^{\mathrm{ref}}+T_z)\times B_z}$.

\paragraph{Distinguishing reference and generated content.}
Reference tokens use positional indices
\begin{equation}
    \operatorname{pos}_{\mathrm{ref}}(i,j)=(1,i,j),
    \qquad j=0,\ldots,T_z^{\mathrm{ref}}-1.
    \label{eq:reference_index}
\end{equation}
The first index ($1$ vs.\ $0$ in Equation~\ref{eq:shared_index}) distinguishes
reference tokens from main-trajectory and music tokens. Reference and main time
indices both start at zero within their respective regions, so prepending the
reference does not offset the main sequence's temporal alignment with music.

\subsection{Training and Inference}
\label{sec:flow}
\paragraph{Training.}
After training the trajectory-image VAE, we freeze it and the music feature
extractor and train the flow Transformer using flow matching
\citep{lipman2022flowmatching}. Given a target latent $Z_{\mathrm{main}}$,
noise $Z_0\sim\mathcal N(0,I)$, and flow time $\tau\in[0,1]$, the Transformer
predicts $\hat U_\theta=f_\theta^{\mathrm{main}}(\widetilde Z_\tau,\tau,A)$.
The interpolation path and flow-matching loss are
\begin{align}
    Z_\tau&=(1-\tau)Z_0+\tau Z_{\mathrm{main}},
    \label{eq:flow_path}\\
    \mathcal L_{\mathrm{FM}}
      &=\mathbb E\!\left[\left\|\hat U_\theta
          -(Z_{\mathrm{main}}-Z_0)\right\|_F^2\right].
    \label{eq:flow_loss}
\end{align}
Only the main sequence is interpolated and supervised; reference latents
remain clean.

\paragraph{Segment sampling.}
At inference (Figure~\ref{fig:method_overview}, bottom left), numerical
integration of the predicted velocity transports
Gaussian noise from $\tau=0$ to $\tau=1$ under the music and reference
conditions. At every step, the unchanged reference latent is prepended
along the sequence-time axis to the current main state; only the main state
is evolved.
The decoder maps the terminal main latent to trajectory logits, from which
we recover the keypoint sequence and validity indicators as described in
Appendix~\ref*{app:keypoint_recovery}.

\paragraph{Long-track inference.}
For music longer than one segment, we generate keypoint segments sequentially,
conditioning each segment after the first on the last generated frame of its
predecessor as reference keypoints (Appendix~\ref*{app:long_track}).

Final videos are synthesized with SCAIL \citep{yan2026scail}; our training-free
extension using noise rescheduling and dynamic overlapping windows is
described in Appendix~\ref*{sec:rendering}.

\section{Experiments}
\label{sec:experiments}
\label{sec:results}

\subsection{Experimental Setup}
\label{sec:exp_setup}

\paragraph{Data.}
We train the music-conditioned keypoint-sequence generator on 35,390 in-the-wild human dance segments from Bilibili and TikTok. We detect keypoints using Sapiens \citep{khirodkar2024sapiens,khirodkarsapiens2}
and map the detections to 83 two-dimensional keypoints (20 body, 42 hand,
and 21 face), yielding 166 coordinate channels with explicit validity information.
We use 43 distinct character--music pairs for evaluation. The music excerpts
last 13--15 seconds, and the reference images include realistic humans,
anime characters, and anthropomorphic animals. Videos are generated
at 30 frames per second.

\paragraph{Implementation details.}
The flow Transformer is a 26-layer NextDiT \citep{Lumina-image-2.0}. Because we
reframe keypoint-sequence generation as image synthesis
(Section~\ref{sec:reframing}), we keep its Transformer blocks unchanged and only
adapt the patch size, the input, output, and conditioning projections, and the
positional indices (Sections~\ref{sec:alignment} and~\ref{sec:reference}).
Sampling uses 25 steps and classifier-free guidance scale 6. Other settings are in
Appendix~\ref*{app:experimental_details}.

\paragraph{Evaluation.}
Following MusicInfuser \citep{hong2026musicinfuser} and Wan-Dancer
\citep{huang2026wandancer}, we score generated videos with a multimodal LLM.
Whereas MusicInfuser uses VideoLLaMA 2 \citep{cheng2024videollama2}, we use the
more recent Gemini 3.8 Flash for its stronger ability
to jointly assess video, audio, and the reference image. We evaluate dance quality (style
alignment, beat alignment, movement coherence, and choreography
complexity) and visual quality (body representation,
video quality, aesthetic quality, and overall consistency), scoring each
criterion from 0 to 10 (higher is better).
Appendix~\ref*{app:eval_prompts} lists the exact prompts.
We also conduct a user study on an in-house crowdsourcing platform with 47 raters. Every rater evaluates all character--music pairs, choosing for each pair the preferred video between EverybodyDance and a comparison method under the same music.

\begin{table*}[t]
\centering
\caption{Quantitative comparison. Each criterion is scored from 0 to 10 (higher is better). Human and Stylized give the mean of the eight scores over realistic-human and stylized references (Appendix~\ref*{app:results_by_type}). Pref. (\%) is the percentage of user-study judgments favoring EverybodyDance over the listed method. $^{\dagger}$ denotes videos generated from X-Dancer's first-stage output using the same SCAIL backbone and inference settings as EverybodyDance.}
\label{tab:main_comparison}
\footnotesize
\setlength{\tabcolsep}{4pt}
\begin{tabular*}{\textwidth}{@{\extracolsep{\fill}}lccccccccccc@{}}
\hline
& \multicolumn{8}{c}{All pairs} & \multicolumn{2}{c}{Mean by type} & \\
\cline{2-9}\cline{10-11}
Method & Style & Beat & Motion & Choreo. & Body
       & Video & Aesthetic & Consist. & Human & Stylized & Pref. (\%) \\
\hline
X-Dancer
& 5.94 & 5.74 & 5.00 & 4.89 & 4.80 & 5.04 & 5.42 & 4.95 & 5.86 & 4.81 & 99.11 \\
MusicInfuser
& 6.24 & 6.22 & 6.06 & 5.54 & 3.49 & 3.15 & 2.97 & 3.05 & 4.68 & 4.53 & 95.70 \\
Wan-Dancer
& 6.20 & 6.14 & 5.97 & 5.39 & 5.61 & 5.65 & 5.71 & 5.53 & 6.46 & 5.33 & 97.48 \\
X-Dancer$^{\dagger}$
& 6.36 & 6.25 & 6.08 & 5.27 & 6.20 & 6.60 & 6.71 & 6.37 & 6.56 & 6.02 & 97.23 \\
Ours
& \textbf{6.50} & \textbf{6.45} & \textbf{6.33} & \textbf{5.74}
& \textbf{6.43} & \textbf{6.71} & \textbf{6.81} & \textbf{6.62}
& \textbf{6.69} & \textbf{6.29} & -- \\
\hline
\end{tabular*}
\end{table*}

\subsection{Quantitative and Qualitative Comparison}
\label{sec:comparison}

Table~\ref{tab:main_comparison} compares EverybodyDance with
X-Dancer \citep{chen2025xdancer}, MusicInfuser \citep{hong2026musicinfuser},
Wan-Dancer \citep{huang2026wandancer}, and X-Dancer$^{\dagger}$.
The last variant uses X-Dancer's first-stage output to condition video
generation with the same SCAIL backbone and inference settings as EverybodyDance.
Our method achieves the highest scores across all eight criteria. The pairwise user study also strongly favors
EverybodyDance, with preference rates of 95.70\%--99.11\% across all baselines.
Side-by-side supplementary videos with audio (Videos~S3 and~S4) complement these scores by
showing motion continuity and music synchronization across reference
characters and music clips.

MusicInfuser generates videos from music and text without a reference-image
input. It achieves a competitive movement coherence score of 6.06, but lower
video quality (3.15) and overall consistency (3.05).
In Video~S3, it produces plausible human dance
motion but does not reproduce the target character or scene appearance.

\begin{figure*}[t]
    \centering
    \includegraphics[width=\textwidth]{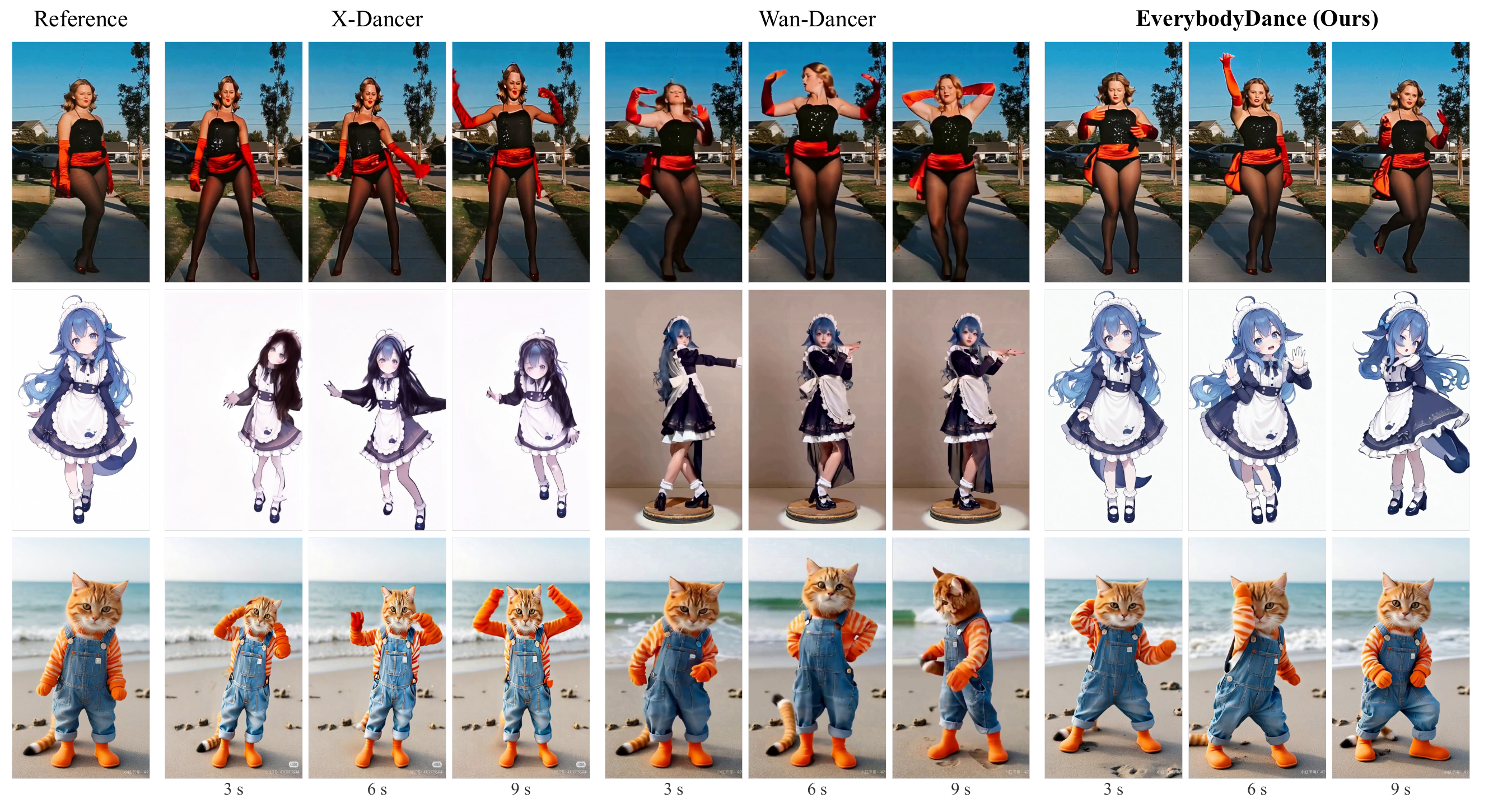}
    \caption{Qualitative comparison across character types.
    Each row uses the same reference image and music across methods,
    with frames shown at 3, 6, and 9 seconds.
    MusicInfuser, which takes no reference image, is omitted here and
    included in Video~S3.}
    \label{fig:qualitative_comparison}
\end{figure*}

Among the native reference-conditioned baselines, Wan-Dancer has the strongest
visual scores, yet EverybodyDance scores higher in video quality
(6.71 vs.\ 5.65) and overall consistency (6.62 vs.\ 5.53).
Wan-Dancer learns music-conditioned synthesis within a global-to-local video
generation framework, rather than generating an explicit keypoint sequence
for a separate pose-conditioned model \citep{huang2026wandancer}.
Figure~\ref{fig:qualitative_comparison} complements the aggregate scores with
examples of a realistic human, an anime-style character, and an anthropomorphic
animal. X-Dancer and Wan-Dancer exhibit appearance and motion artifacts on the
tested anime and anthropomorphic references, while EverybodyDance better
preserves the reference appearance and produces more convincing dance motions.
Video~S3 further shows color shifts, distorted proportions,
and loss of facial or garment details in X-Dancer and Wan-Dancer on
stylized, anime, line-art, and anthropomorphic examples. Some outputs drift
toward realistic human shapes, whereas EverybodyDance better retains the reference character.
Our modular design allows music--dance relationships to be learned
from real-human keypoint sequences while a pretrained video generation model
handles cross-style appearance synthesis. Splitting the evaluation pairs by
character type supports this quantitatively (Table~\ref{tab:main_comparison}):
EverybodyDance has the highest mean score for both realistic humans (6.69) and
stylized characters (6.29), and its lead over Wan-Dancer grows from 0.23 to
0.96, as does user preference (95.49\% to 98.77\%;
Appendix~\ref*{app:results_by_type}).

Replacing X-Dancer's native second-stage model with SCAIL, conditioned on
X-Dancer's first-stage output as explicit keypoints, raises its video quality
score from 5.04 to 6.60. An explicit keypoint interface thus lets a
music-driven system use a stronger pose-conditioned video generation model.
With SCAIL and its inference settings held fixed, EverybodyDance still scores
higher than X-Dancer$^{\dagger}$ in
movement coherence (6.33 vs.\ 6.08), choreography complexity (5.74 vs.\ 5.27),
and overall consistency (6.62 vs.\ 6.37), and receives 97.23\% user preference.
Video~S4 shows improved appearance preservation
for X-Dancer$^{\dagger}$, but also stiffer torso movement, less precise beat
synchronization, and intermittent limb artifacts compared with EverybodyDance.
Since both use the same second stage, the remaining
differences come from the generated keypoint sequences, supporting the
effectiveness of our first-stage design beyond the choice of second-stage model.

\subsection{Ablation on Trajectory-Image VAE Training}
\label{sec:codec_ablation}

We compare four training configurations for the trajectory-image VAE,
evaluating their 100,000-step checkpoints on the same holdout of 512
sequences with 256 frames each. \emph{Global} denotes hand and facial
landmarks expressed in image coordinates, rather than relative to the
wrist or nose. \emph{Global MSE} directly minimizes mean squared error
(MSE) between the input trajectory images and the VAE's reconstructed
output maps. \emph{Global CE} replaces the MSE reconstruction loss with
cross-entropy (CE) between predicted spatial-bin distributions and the
Gaussian-smoothed spatial labels defined in Appendix~\ref*{app:vae_training}.
It predicts 512 spatial classes and excludes missing coordinate axes from
the reconstruction loss.
\emph{Global CE + null} adds a null class to the same softmax, yielding
513 output classes, and supervises missing axes through this class.
\emph{Local CE + null} retains the CE and null supervision while keeping body
joints in image coordinates, using wrist-relative hand coordinates with an
asinh warp and nose-relative facial coordinates with linear normalization.
All four configurations use the same KL regularization schedule
(Appendix~\ref*{app:vae_ablation_protocol}), and every metric is computed on
ground-truth masks shared by all four (Appendix~\ref*{app:metric_masks}).

\begin{table}[t]
\centering
\caption{Ablation of VAE training configurations. Overall and body L2 measure
image-space reconstruction error over all 83 landmarks and the 20 body
landmarks; hand and face L2 subtract the corresponding wrist and nose positions
from both predictions and targets.
L2 is reported in $10^{-3}$ normalized image coordinates.
Missing F1 and FNR are percentages with missing landmarks as the positive class.
All configurations share the same ground-truth evaluation masks; see
Appendix~\ref*{app:vae_metrics} for definitions.}
\label{tab:codec_ablation}
\scriptsize
\setlength{\tabcolsep}{2.5pt}
\begin{tabular*}{\linewidth}{@{\extracolsep{\fill}}lcccccc@{}}
\hline
& \multicolumn{4}{c}{L2 $\downarrow$} & \multicolumn{2}{c}{Missing (\%)} \\
\cline{2-5}\cline{6-7}
Training setup & Overall & Body & Hand & Face & F1 $\uparrow$ & FNR $\downarrow$ \\
\hline
Global MSE       & 4.20 & 3.94 & 8.43 & 3.11 & 40.50 & 30.48 \\
Global CE        & 2.95 & 2.88 & 4.26 & 2.30 & 63.10 & 41.99 \\
Global CE + null & \textbf{2.90} & 2.86 & 4.43 & 2.40
                 & \textbf{99.58} & \textbf{0.54} \\
Local CE + null  & 3.86 & \textbf{2.78} & \textbf{2.38} & \textbf{0.51}
                 & 99.51 & 0.70 \\
\hline
\end{tabular*}
\end{table}

Table~\ref{tab:codec_ablation} shows that replacing MSE with CE lowers all
four L2 errors. Missing F1
increases from 40.50\% to 63.10\%, but FNR also increases from
30.48\% to 41.99\%, showing that better coordinate reconstruction
does not ensure reliable missingness prediction. With explicit
null supervision, Missing F1 reaches 99.58\% and FNR falls to
0.54\%, without separate confidence calibration.
Because Table~\ref{tab:codec_ablation} measures codec reconstruction fidelity,
the high Missing F1 shows that the latents retain explicit missingness
information for the flow Transformer to learn. Explicit null supervision thus
enables near-perfect missing-landmark recognition, while local parameterization
achieves the lowest error across all three anatomical groups (Body L2, Hand
aligned L2, and Face aligned L2).

Compared with Global CE + null, the local variant reduces body L2 from 2.86 to
2.78, hand aligned L2 by 46\% (4.43 to 2.38), and face aligned L2 by 79\%
(2.40 to 0.51), while overall L2 increases from 2.90 to 3.86.
This is consistent with anchor-relative reconstruction: hand and facial
landmarks make up 76\% of the 83 keypoints, and an error in a predicted wrist or
nose rigidly shifts all 21 landmarks reconstructed relative to it, without
distorting their shape. We select the local training configuration to prioritize fine hand
articulation in dance while retaining a low Missing FNR of 0.70\%.

\subsection{Ablations on Representation and Temporal Alignment}
\label{sec:control_ablation}

We compare alternative coordinate-based generation designs and a temporal
alignment variant. For a fair comparison, Raw Coordinates, Raw Coordinates +
VAE, and EverybodyDance use the same 26-layer DiT backbone, shared-time RoPE strategy, music-conditioning mechanism, training objective,
optimizer, and flow-matching schedule. Across these variants, only the generator's input and output projections are
adapted to each representation.
All generated keypoint sequences condition the same fixed, pretrained SCAIL
video generation model with identical inference settings.
Appendix~\ref*{app:generation_variants} gives the architecture details.

\paragraph{Coordinate-based generation.}
\emph{Raw Coordinates} directly generates continuous 2D keypoint coordinates
conditioned on music as a width-one ``thin image,'' without spatial-bin
encoding or an autoencoder.
\emph{Raw Coordinates + VAE} uses a 1D temporal-convolutional encoder--decoder
adapted from Bailando \citep{Bailando2}, replacing its VQ codebook with a
continuous diagonal Gaussian posterior of comparable parameter count to our
trajectory-image VAE. The DiT generates continuous latents that the VAE
decodes back into keypoint sequences (Appendix~\ref*{app:generation_variants}).

\paragraph{Music--trajectory temporal alignment.}
\emph{T2I RoPE} retains our trajectory-image VAE and reference trajectory
conditioning but replaces the shared temporal indexing in
Section~\ref{sec:alignment}. Instead of assigning corresponding music and
trajectory tokens the same temporal index, it gives the music token at latent
time $j$ the position $(j+2,0,0)$: music tokens take independent sequential
positions on the first RoPE axis, as in text-to-image models. The offset of two
skips the first-axis values $\{0,1\}$ used by the main and reference tokens.
The trajectory and reference token positions remain unchanged.

\begin{table}[t]
\centering
\caption{Representation and temporal-alignment variants.
Raw abbreviates Raw Coordinates.
Pref. (\%) is the percentage of user-study judgments favoring EverybodyDance
over the listed variant.}
\label{tab:control_ablation}
\scriptsize
\setlength{\tabcolsep}{1.5pt}
\begin{tabular*}{\linewidth}{@{\extracolsep{\fill}}lccccccccc@{}}
\hline
Variant & Style & Beat & Motion & Choreo. & Body
        & Video & Aesthetic & Consist. & Pref. (\%) \\
\hline
Raw
& 6.34 & 6.25 & 5.77 & 5.60 & 5.85 & 6.32 & 6.45 & 5.93 & 96.83 \\
\shortstack[l]{Raw\\+ VAE}
& 6.36 & 6.24 & 5.99 & 5.71 & 6.03 & 6.33 & 6.51 & 6.12 & 90.10 \\
T2I RoPE
& 6.43 & 6.35 & 6.18 & 5.57 & 6.29 & 6.54 & 6.69 & 6.47 & 86.69 \\
Ours
& \textbf{6.50} & \textbf{6.45} & \textbf{6.33} & \textbf{5.74}
& \textbf{6.43} & \textbf{6.71} & \textbf{6.81} & \textbf{6.62} & -- \\
\hline
\end{tabular*}
\end{table}

Table~\ref{tab:control_ablation} shows that EverybodyDance achieves the
highest scores on all eight video-level criteria.
For the coordinate-based designs, movement coherence improves from 5.77
with Raw Coordinates to 5.99 with Raw Coordinates + VAE, but remains below
our score of 6.33. EverybodyDance is preferred over these two variants in
96.83\% and 90.10\% of user-study judgments, respectively.
In Video~S5, Raw Coordinates produces implausible body poses, unnatural limb
geometry, and structural distortion over time. Raw Coordinates + VAE exhibits
abrupt changes in character scale and floating feet, while EverybodyDance
maintains more coherent body structure and motion.

For temporal alignment, our shared indexing scores higher than T2I RoPE
in beat alignment (6.45 vs.\ 6.35) and movement coherence (6.33 vs.\ 6.18),
with 86.69\% user preference. In Video~S6, T2I RoPE preserves
body proportions and foot contact, but its motion
transitions align less precisely with musical beats and phrasing than those
of EverybodyDance.

\section{Conclusion}
\label{sec:conclusion}

We presented EverybodyDance, which generates character dance videos from music
and a single reference image by reformulating 2D keypoint-sequence generation
as music-conditioned trajectory-image synthesis and passing the generated
sequences to a pretrained pose-conditioned video generation model.
\textbf{Limitations.} Our 2D keypoint interface assumes a humanoid body, so
characters without a human-like skeleton are outside its current scope. The
system also animates a single character, so multi-person and group dance are
not yet supported. Since trajectory images can carry additional channels,
future work could support multiple dancers by adding keypoint channels for each
person and training on multi-person dance data.

{
    \small
    \bibliographystyle{ieeenat_fullname}
    \bibliography{main}
}

\maketitlesupplementary
\section*{Supplementary Videos}
The main paper refers to the following videos, provided with this document.
Each file name starts with the video ID.
\begin{description}
  \item[Video S1] EverybodyDance results for realistic humans, anime and
    line-art characters.
  \item[Video S2] A dance result using a template reference skeleton
    (Section~\ref*{sec:reference} and Appendix~\ref{app:template_reference}).
  \item[Video S3] Side-by-side comparison with X-Dancer, MusicInfuser, and
    Wan-Dancer (Section~\ref*{sec:comparison}).
  \item[Video S4] Comparison with X-Dancer$^{\dagger}$, which uses the same
    SCAIL model and inference settings as EverybodyDance
    (Section~\ref*{sec:comparison}).
  \item[Video S5] Ablation of coordinate-based generation: Raw Coordinates and
    Raw Coordinates + VAE (Section~\ref*{sec:control_ablation}).
  \item[Video S6] Ablation of temporal alignment: T2I RoPE
    (Section~\ref*{sec:control_ablation}).
\end{description}

\appendix
\counterwithin{table}{section}
\counterwithin{equation}{section}

\section{Fine-Grained Keypoint Parameterization}
\label{app:finegrained}

We use part-relative coordinates for hand and facial landmarks before applying
the categorical encoding of Section~\ref*{sec:reframing}. Here, a local
``crop'' refers to a transformation of keypoint coordinates, not to cropping
hand or head images from the video. Body keypoints remain in normalized image
coordinates. Hand and facial landmarks are instead expressed relative to a
body anchor and mapped into $[0,1]$, where valid axes are encoded as one-hot
vectors over the same $B=512$ spatial bins. These vectors are stacked
across channels and time as described in Section~\ref*{sec:reframing}.
Image-level preprocessing is specified in Appendix~\ref{app:input_preprocessing}.
Missing axes use zero vectors; the null category appears only in the
decoder output for missing-coordinate supervision. The two parts use different
transformations: an asinh warp for hands and a linear map for the face.

\subsection{Wrist-Relative Hand Coordinates}
\label{app:hand_coordinates}

We process the left and right hands separately. Let
$\mathbf p_{h,k,t}\in[0,1]^2$ be the normalized full-image coordinates of
hand landmark $k$ at frame $t$, and let $\mathbf p_{w,t}$ be the same-side
wrist position. We first compute the wrist-relative displacement:
\begin{equation}
    \boldsymbol\delta_{h,k,t}
      = \mathbf p_{h,k,t}-\mathbf p_{w,t}.
    \label{eq:hand_relative}
\end{equation}
Each horizontal or vertical component, denoted by $\Delta$, is transformed
independently using an inverse-hyperbolic-sine (\emph{asinh}) warp.
The function $\operatorname{asinh}(z)=\ln(z+\sqrt{1+z^2})$ provides a
sign-preserving nonlinear compression of displacement.

\paragraph{Forward warp.}
We first map the displacement to $[-1,1]$ and then shift and scale it
to $[0,1]$:
\begin{equation}
    \begin{aligned}
        w(\Delta)
          &=\operatorname{sgn}(\Delta)\,
            \frac{\operatorname{asinh}\!\left(
                  \min(|\Delta|,R_{\mathrm{cap}})/s\right)}
                 {\operatorname{asinh}(R_{\mathrm{cap}}/s)},\\
        u_h(\Delta)&=\frac{w(\Delta)+1}{2}.
    \end{aligned}
    \label{eq:hand_warp}
\end{equation}
Because asinh is odd, the first line simplifies to
$w(\Delta)=\operatorname{asinh}(\Delta/s)/
\operatorname{asinh}(R_{\mathrm{cap}}/s)$ whenever
$|\Delta|\le R_{\mathrm{cap}}$.
Zero displacement maps to $u_h=0.5$, and the endpoints
$\pm R_{\mathrm{cap}}$ map to $0$ and $1$.
Valid transformed coordinates are quantized into the $B=512$ spatial
bins and stacked into trajectory images as described in
Section~\ref*{sec:reframing}.
Missing axes remain unavailable and use zero coordinate encodings;
they are not assigned the wrist position.

\paragraph{Dataset-level parameters.}
Both parameters are computed over the full training set and held fixed
thereafter. The scale $s\simeq0.0602$ is the median ($p_{50}$) of the
hand-to-wrist span statistic and controls the transition from approximately
linear behavior near the wrist to compression at larger displacements.
The cap $R_{\mathrm{cap}}\simeq0.5496$ is the maximum observed hand-to-wrist
offset, covering the training-set displacement range, including close-up
hands.

\paragraph{Inverse warp.}
Given a decoded local coordinate $\hat u_h\in[0,1]$, we restore the
normalized signed displacement and invert the nonlinear map:
\begin{equation}
    \begin{aligned}
        \hat w&=2\hat u_h-1,\\
        \hat\Delta
          &=s\,\sinh\!\left(\hat w\,
                     \operatorname{asinh}(R_{\mathrm{cap}}/s)\right),\\
        \hat{\mathbf p}_{h,k,t}
          &=\hat{\mathbf p}_{w,t}
            +\hat{\boldsymbol\delta}_{h,k,t}.
    \end{aligned}
    \label{eq:hand_inverse}
\end{equation}
The scalar inverse is applied to both axes to form
$\hat{\boldsymbol\delta}_{h,k,t}$, after which the predicted wrist position
is added back. Since $\sinh$ is odd, this expression is equivalent to
multiplying $s\,\sinh(|\hat w|\operatorname{asinh}(R_{\mathrm{cap}}/s))$
by $\operatorname{sgn}(\hat w)$.
The anchor $\hat{\mathbf p}_{w,t}$ is reconstructed or generated, not supplied
from the ground truth at inference.
Before quantization, the inverse exactly recovers displacements within
$[-R_{\mathrm{cap}},R_{\mathrm{cap}}]$; beyond this range it recovers the
clipped displacement.

\paragraph{Why use asinh instead of a linear map?}
Typical full- or half-body views contain much smaller hand displacements
than occasional close-up views. A linear map over the entire capped range
would assign only a narrow central portion of the 512 bins to the small
finger movements common in these views.
For $|\Delta|\ll s$, $\operatorname{asinh}(\Delta/s)$ is approximately
linear in $\Delta$; for $|\Delta|\gg s$, its magnitude grows approximately
logarithmically. The normalized warp therefore allocates more bins per
unit displacement near the wrist and fewer at large offsets, while
retaining the full capped range. This balances local coordinate resolution
with coverage of large hands rather than shrinking the representation range.

\subsection{Nose-Relative Facial Coordinates}
\label{app:face_coordinates}

For the $21$ facial landmarks, we use the nose tip as the anchor.
Given landmark coordinates $\mathbf p_{f,k,t}$ and nose coordinates
$\mathbf p_{\mathrm{nose},t}$, we define
\begin{equation}
    \begin{aligned}
        \boldsymbol\delta_{f,k,t}
          &=\mathbf p_{f,k,t}-\mathbf p_{\mathrm{nose},t},\\
        \mathbf u_{f,k,t}
          &=\frac{\boldsymbol\delta_{f,k,t}+R}{2R},
          \qquad R\simeq0.0698,
    \end{aligned}
    \label{eq:face_linear}
\end{equation}
with scalar operations applied to each axis. The nose maps to
$(0.5,0.5)$ and the displacement window $[-R,R]$ maps linearly to $[0,1]$.
Unlike hand coordinates, facial coordinates are \emph{not} asinh-warped.
For valid decoded facial coordinates, the image-space reconstruction is
\begin{equation}
    \hat{\mathbf p}_{f,k,t}
      =\hat{\mathbf p}_{\mathrm{nose},t}
        +R\left(2\hat{\mathbf u}_{f,k,t}-1\right).
    \label{eq:face_inverse}
\end{equation}

\paragraph{Segment-level facial validity.}
The face window is handled differently from the hand clipping range.
If any originally valid facial landmark in any frame of a segment lies
outside $[-R,R]$ on either coordinate axis, we mark the \emph{entire facial
modality of that segment} as invalid. Writing $\mathcal V_f$ for the
originally valid facial landmark--frame pairs, this segment flag is
\begin{equation}
    q_f =
    \mathbf 1\!\left[
    \begin{gathered}
        \left|\delta^a_{f,k,t}\right|\le R\\
        \forall\,(k,t)\in\mathcal V_f,\quad a\in\{x,y\}
    \end{gathered}
    \right].
    \label{eq:face_segment_validity}
\end{equation}
When $q_f=0$, the facial targets of this segment are marked as missing, while
its body and hand targets are kept. Together with the requirement of a valid
nose anchor, this check retains
99.83\% of the originally valid facial landmark instances in the full dataset.

\subsection{Categorical Encoding and Reconstruction}
\label{app:finegrained_reconstruction}

For hands and faces, the transformed coordinates $\mathbf u_h$ and
$\mathbf u_f$, rather than their absolute image coordinates, are supplied
to the spatial-bin encoding. The decoded bin location is a local coordinate
$\hat u$; Equations~\ref{eq:hand_inverse} and~\ref{eq:face_inverse} restore
its image-space position using the corresponding predicted anchor.
Valid local-coordinate targets use the Gaussian spatial labels described
in Appendix~\ref{app:vae_training}. Coordinate validity is tracked separately from
these geometric transformations. Missing axes are encoded as all-zero
vectors before stacking and supervised through the decoder's null output;
a zero vector
does not assign them a spatial displacement of zero.

\paragraph{Reconstruction comparison.}
Table~\ref*{tab:codec_ablation} (Section~\ref*{sec:codec_ablation}) compares
Local CE + null, which uses this part-relative parameterization, with
Global CE + null, which keeps hand and facial landmarks in image coordinates.
The local configuration reduces body L2 from 2.86 to 2.78, hand aligned L2
from 4.43 to 2.38, and face aligned L2 from 2.40 to 0.51 (in $10^{-3}$
normalized image coordinates), but increases overall L2 from 2.90 to 3.86. This
is consistent with anchor-relative reconstruction: an error in a predicted wrist
or nose rigidly shifts all landmarks reconstructed relative to it, without
distorting their shape.

\section{Recovering Keypoint Sequences from Trajectory Images}
\label{app:keypoint_recovery}

This section details how the trajectory-image representation is decoded into
a sequence of keypoint coordinates and validity indicators. Let $K$ be the
number of keypoints, $T$ the sequence length, $B=512$ the number of spatial
bins, and $C=2K$ the number of coordinate channels. The channel indices are
$c_x(k)=2k-1$ and $c_y(k)=2k$, following Section~\ref*{sec:reframing}.

The VAE decoder produces logits that we transpose back to the trajectory-image
layout, $\hat L\in\mathbb R^{C\times(B+1)\times T}$,
with $B$ spatial logits and one null logit per coordinate axis. The null
category, indexed by $B$, is supervised by the missing-coordinate loss
described in Appendix~\ref{app:vae_training}.
For each keypoint and axis, we read the most likely output category and its
validity:
\begin{equation}
    \begin{aligned}
        \hat b^a_{k,t}
          &=\operatorname*{arg\,max}_{b\in\{0,\ldots,B\}}
             \hat L[c_a(k),b,t],\\
        \hat v^a_{k,t}&=\mathbf 1[\hat b^a_{k,t}<B].
    \end{aligned}
    \label{eq:decode_category}
\end{equation}
Using the lower-edge bin convention, the corresponding coordinate is
\begin{equation}
    \hat a_{k,t}=
    \begin{cases}
        \hat b^a_{k,t}/B, & \hat v^a_{k,t}=1,\\
        \bot, & \hat v^a_{k,t}=0,
    \end{cases}
    \qquad a\in\{x,y\},
    \label{eq:decode_coordinate}
\end{equation}
where $\bot$ denotes a missing coordinate, not the numeric value zero.
For hand and facial landmarks, we apply the inverse transformations in Appendix~\ref*{app:finegrained}
to recover image coordinates before passing the generated keypoint sequence
to the second-stage video model.

\section{Trajectory-Image VAE Training Details}
\label{app:vae_training}

This appendix gives the reconstruction targets, losses, and
training objective for the trajectory-image VAE introduced in
Section~\ref*{sec:codec}. The coordinate and time indices follow
Section~\ref*{sec:reframing}.

\subsection{Coordinate and Missingness Targets}
For each coordinate axis at each frame, the decoder predicts a distribution
over $B=512$ spatial bins and one null category. Valid coordinates receive
spatial supervision, while missing coordinates receive the null label.
The resulting reconstruction problem has 513 output classes.
Let $i=(n,k,t,a)$ identify sample $n$, keypoint $k$, frame $t$, and coordinate
axis $a\in\{x,y\}$. Writing the corresponding logits as $\ell_{i,b}$, we compute
\begin{equation}
    p_{i,b}
      = \frac{\exp\ell_{i,b}}{\sum_{j=0}^{512}\exp\ell_{i,j}},
      \qquad b=0,\ldots,512.
    \label{eq:codec_softmax}
\end{equation}
The spatial and null categories share this softmax, so coordinate validity
is predicted together with location rather than by a separate classifier.
For a valid axis, we use Gaussian-smoothed reconstruction targets $g_{i,b}$
centered near the ground-truth coordinate over the spatial bins (standard
deviation of 6 bins), with zero target weight on null.
For a missing axis, supervision selects only the null category and assigns
no target spatial location.

\subsection{Reconstruction Loss}
Let $\mathcal I_{\mathrm v}$ and $\mathcal I_{\mathrm m}$ denote the valid
and missing coordinate axes in a training batch, as determined by the
annotation validity indicators. We compute separate mean cross-entropy
losses for the two groups:
\begin{equation}
    \begin{aligned}
        \mathcal L_{\mathrm{valid}}
          &= -\frac{1}{|\mathcal I_{\mathrm v}|+\epsilon}
             \sum_{i\in\mathcal I_{\mathrm v}}\sum_{b=0}^{511}
             g_{i,b}\log p_{i,b},\\
        \mathcal L_{\mathrm{missing}}
          &= -\frac{1}{|\mathcal I_{\mathrm m}|+\epsilon}
             \sum_{i\in\mathcal I_{\mathrm m}}\log p_{i,512},\\
        \mathcal L_{\mathrm{rec}}
          &= \frac{\mathcal L_{\mathrm{valid}}
                     +0.5\,\mathcal L_{\mathrm{missing}}}{\ln 513}.
    \end{aligned}
    \label{eq:codec_ce}
\end{equation}
Here $\epsilon=10^{-6}$ prevents division by zero when a group is empty.
Separate averaging controls the relative contribution of valid and missing
axes independently of their counts; the missing-coordinate term is weighted
by $0.5$, and $\ln 513$ rescales the combined loss. Because all categories
share the same softmax, assigning probability to null for a valid axis also
reduces the probability available to its spatial target. Conversely, a
missing axis is supervised without inventing a coordinate.

\subsection{VAE Training Objective}
We combine categorical reconstruction and KL regularization of the latent
posterior:
\begin{equation}
    \mathcal L_{\mathrm{codec}}
      = \mathcal L_{\mathrm{rec}}
        +\beta(s)\,\mathcal L_{\mathrm{KL}}.
    \label{eq:codec_total}
\end{equation}
The KL term regularizes the posterior toward a standard Gaussian, with
$\beta(s)$ controlling its weight during warmup at VAE training step $s$.
After VAE training, we freeze both the encoder and decoder. The
music-conditioned generator then learns to synthesize trajectory latents;
the fixed decoder converts them into keypoint coordinates and validity
indicators using the procedure in Appendix~\ref*{app:keypoint_recovery}.

\section{Additional Experimental Details}
\label{app:experimental_details}

\subsection{Training and Sampling Settings}
\label{app:training_settings}

We sample 256-frame windows with starts aligned to multiples of eight frames
and pair each window with 32 temporally aligned, 4,800-dimensional music
features, extracted and aligned as described in Section~\ref*{sec:alignment}.
A 256-frame, 166-channel trajectory-image input is encoded into a
$72\times32\times64$ latent, ordered as channels, time, and spatial position,
and repeating the reference keypoints for $T_{\mathrm{ref}}=32$ frames yields a
clean $72\times4\times64$ reference latent. We use the codec checkpoint at
100,000 steps, trained with AdamW on eight GPUs (global batch size 64; learning
rate $10^{-4}$), and freeze it during generator training. The
music-conditioned generator uses a 26-layer NextDiT backbone with hidden
dimension 2,304, 24 query heads, eight key/value heads, and patch size one.
The coordinate-based variants use the same backbone
(Appendix~\ref{app:generation_variants}). We train the generator on eight GPUs
with AdamW (global batch size 128; learning rate $2\times10^{-4}$), bfloat16
forward computation, and 25\% audio-condition dropout. We use the 66,000-step
checkpoints, 25 sampling steps, and classifier-free guidance scale 6.
For second-stage video synthesis, X-Dancer$^{\dagger}$, EverybodyDance, and
all keypoint-sequence generation variants in Section~\ref*{sec:control_ablation}
use the same pretrained SCAIL model and inference settings
(Appendix~\ref{sec:rendering}).

\subsection{Input Preprocessing and Music Feature Alignment}
\label{app:input_preprocessing}

We first stack the 512-bin one-hot coordinate encodings, with zero vectors for
missing axes, across channels and time to construct the trajectory image $H(X)$
in Section~\ref*{sec:reframing}. We then convolve each one-hot vector along the spatial-bin axis with the
normalized Gaussian kernel $k(d)=\exp\bigl(-d^2/(2\sigma^2)\bigr)/(\sigma\sqrt{2\pi})$,
where $d$ is the bin offset and $\sigma=6$ bins, and denote the result by
$\tilde H(X)$. The convolution acts within each frame and never across time,
so the zero vectors of missing axes remain zero. The same smoothed vectors
serve as the valid-axis reconstruction targets $g_{i,b}$
(Appendix~\ref{app:vae_training}). Before encoding, $\tilde H(X)$ is
transposed to channels $\times$ time $\times$ spatial bins (the Transpose step
in Figure~\ref*{fig:method_overview}), and the latents keep this order. The
main-paper notation $E_{\phi}(H(X))$ includes this preprocessing in the
encoding operation. The decoder predicts 513 classes, adding a null output for
missing-coordinate supervision (Appendix~\ref{app:vae_training}).

Using JukeMIR \citep{castellon2021codified}, we extract music features
$\Psi(m)$ from layer 66 (level 2) of the frozen Jukebox 5B model
\citep{dhariwal2020jukebox}. For a keypoint
sequence at $r$ frames per second, the VAE's temporal stride of eight yields
trajectory latents at $r/8$ Hz. We use FFT-based resampling to match the
music-feature rate to this latent temporal rate. For $T$-frame segments, the
resulting music sequence has $T_z=T/8$ tokens, $A\in\mathbb R^{T_z\times d_a}$,
with one music token aligned to each trajectory-latent time step.

\subsection{VAE Ablation Training Protocol}
\label{app:vae_ablation_protocol}

In the VAE ablation study (Section~\ref*{sec:codec_ablation}), Global CE and
Global CE + null share the same base training settings and valid-axis CE
implementation; the latter adds the null output and missing-axis supervision.
All four configurations use the same KL regularization weight and warmup
schedule. For Global MSE and Global CE,
which have no null output, missingness thresholds are selected to maximize
Missing F1 on separate validation sequences and held fixed during evaluation.
All four configurations are evaluated on the same ground-truth masks
(Appendix~\ref{app:metric_masks}).

\subsection{Evaluation Prompts}
\label{app:eval_prompts}

This section gives the exact prompt used with Gemini 3.8 Flash to score the
eight criteria in Section~\ref*{sec:exp_setup}. The prompt is reproduced
verbatim below.

\begin{lstlisting}[
  basicstyle=\fontencoding{T1}\ttfamily\footnotesize\selectfont,
  breaklines=true,
  breakatwhitespace=false,
  columns=fullflexible,
  keepspaces=true,
  showstringspaces=false,
  upquote=true,
  literate={–}{{{\rmfamily\textendash}}}1
]
Evaluate the attached video on eight independent dimensions. Inspect the full video over time and listen to its audio. Judge only evidence in the supplied inputs; do not assume an intended choreography, reference video, or external music. The attached image, when provided, is the reference image for this video. For all appearance-related dimensions (video_quality, aesthetic_quality, overall_consistency, and dancer_body_representation), compare the video against the reference image: preserve its character identity, clothing, proportions, colors, details, visual style, and relevant background. Do not reward a video merely for being sharper or prettier if it departs from the reference. The image is not a reference for motion, choreography, music, or an exact pose or camera view. For dance_music_style_alignment, dance_music_beat_alignment, dancer_movement_coherence, and choreography_complexity, judge the audible music and observable movement independently of the reference image's and video's visual style. Appearance and setting must neither raise nor lower these four scores by themselves. If no reference image is supplied, judge visual qualities from the video alone.

Use a 0–10 scale for every dimension. Decimal scores with at most one decimal place are allowed. Use the full scale when justified. Judge each dimension by its own criteria; do not mechanically copy one score into another.

1. video_quality: Preservation of the reference image's visible detail and rendering quality throughout the video, including sharpness, exposure, noise, compression artifacts, dropped or duplicated frames, and unwanted loss or invention of details. A technically crisp video that changes the reference's character or style should not receive a high score. Without a reference, assess the video's technical visual quality on its own. 0 = poor video quality; 5 = moderate video quality; 10 = excellent video quality with strong reference fidelity when applicable. Do not score dance skill or soundtrack quality here.
2. aesthetic_quality: Fidelity to the reference image's visual style, composition, color harmony, lighting, and presentation as the dancer moves. Do not score standalone prettiness or prefer a different, more polished style over the reference. Without a reference, assess the video's aesthetic quality on its own. 0 = poor aesthetic quality or fidelity; 5 = moderate aesthetic quality or fidelity; 10 = excellent aesthetic quality or fidelity.
3. overall_consistency: Temporal coherence throughout the clip and fidelity to the reference image: stable dancer identity, facial features, hairstyle, clothing, body appearance, distinctive colors, background, and objects; minimal flickering, morphing, discontinuities, or unexplained changes. Without a reference, assess temporal consistency alone. Do not require the same pose, expression, viewpoint, or framing as the reference, and do not treat intentional movement or camera motion as inconsistency. 0 = poor consistency; 5 = moderate consistency; 10 = excellent temporal consistency and reference fidelity when applicable.
4. dance_music_style_alignment: How well the movement vocabulary, mood, energy, and phrasing suit the audible music's style and changes over time. Judge the style of the movement relative to the music, not the visual art style of the dancer or reference image. Appearance and setting are not evidence of musical fit. 0 = poor style alignment; 5 = moderate style alignment; 10 = perfect style alignment. This is stylistic compatibility, separate from precise beat timing.
5. dance_music_beat_alignment: How accurately visible movement accents, steps, direction changes, and pauses align with audible beats, rhythmic accents, and musical phrases over time. Judge only audible-to-visible timing, independent of appearance or setting. Continuous dancing while music plays is not proof of synchronization. 0 = poor beat alignment; 5 = moderate beat alignment; 10 = perfect beat alignment. Account for intentional syncopation rather than demanding every movement land on a main beat.
6. dancer_body_representation: Structural integrity of the dancer across frames, including proportions, limbs, joints, hands, and feet. When a reference image is supplied, compare the character's proportions and stylized anatomy with that image rather than imposing a different design or anatomy; penalize unintended distortions or lost features. Without a reference, assess anatomical plausibility from the video alone. 0 = severely distorted or unfaithful representation; 5 = moderate issues; 10 = faithful and structurally sound representation. Do not rate attractiveness or body type. Do not penalize ordinary occlusion by itself.
7. dancer_movement_coherence: Coherence of the motion itself over time: joint articulation where applicable, balance, weight transfer, momentum, and foot-ground contact. Allow intentional stylization or exaggeration while judging whether movement remains internally coherent. Penalize unintended jitter, sliding feet, limb popping, or abrupt displacements. 0 = unstable or incoherent motion; 5 = moderate issues; 10 = coherent, well-controlled motion. Distinguish intentional movement from temporal artifacts.
8. choreography_complexity: Variety and difficulty of the visible choreography within the audible music's style and rhythm. Reward distinct coordinated phrases, footwork where applicable, and purposeful transitions only when they suit the music. More moves or faster movement alone do not earn a higher score; busy choreography that clashes with the music should not score high. Repeated phrases, visual design, and camera motion do not add complexity. 0 = basic or disconnected choreography; 5 = moderately varied and musically suited; 10 = rich, coordinated choreography that fits the music throughout.

Return exactly one valid JSON object with the following eight keys, each mapped to a number from 0 to 10 or null. No Markdown, explanation, extra keys, or overall/aggregate score:
{"video_quality":0,"aesthetic_quality":0,"overall_consistency":0,"dance_music_style_alignment":0,"dance_music_beat_alignment":0,"dancer_body_representation":0,"dancer_movement_coherence":0,"choreography_complexity":0}
\end{lstlisting}

\subsection{Results by Character Type}
\label{app:results_by_type}

Table~\ref{tab:results_by_type} splits the scores of
Table~\ref*{tab:main_comparison} by reference type: 17 realistic humans and 26
stylized characters, which cover all non-human references, including anime
characters, anthropomorphic animals, and line drawings. EverybodyDance has the
highest mean score in both groups and the highest score on seven of the eight
criteria in each group.

Table~\ref{tab:pref_by_type} splits the user-study preferences in the same
way. Against every baseline, EverybodyDance is preferred in at least 94.11\%
of judgments in each group.

\begin{table*}[t]
\centering
\caption{Video-level scores by reference type. Mean is the average of the eight
scores; bold marks the best score in each group. $^{\dagger}$ is defined as in
Table~\ref*{tab:main_comparison}.}
\label{tab:results_by_type}
\footnotesize
\begin{tabular*}{\textwidth}{@{\extracolsep{\fill}}lccccccccc@{}}
\hline
Method & Style & Beat & Motion & Choreo. & Body
       & Video & Aesthetic & Consist. & Mean \\
\hline
\multicolumn{10}{@{}l}{\emph{Realistic humans (17 pairs)}} \\
X-Dancer             & 6.20 & 6.04 & 5.77 & 5.13 & 5.70 & 5.82 & 6.22 & 5.97 & 5.86 \\
MusicInfuser         & 6.25 & 6.24 & 5.96 & 5.57 & 3.96 & 3.27 & 3.06 & 3.09 & 4.68 \\
Wan-Dancer           & 6.64 & 6.55 & 6.54 & \textbf{5.82} & 6.57 & 6.32 & 6.57 & 6.63 & 6.46 \\
X-Dancer$^{\dagger}$ & 6.66 & 6.61 & 6.62 & 5.60 & 6.68 & 6.75 & 6.83 & 6.69 & 6.56 \\
Ours & \textbf{6.68} & \textbf{6.71} & \textbf{6.85} & 5.77 & \textbf{6.91}
     & \textbf{6.83} & \textbf{6.84} & \textbf{6.96} & \textbf{6.69} \\
\hline
\multicolumn{10}{@{}l}{\emph{Stylized characters (26 pairs)}} \\
X-Dancer             & 5.78 & 5.55 & 4.50 & 4.74 & 4.21 & 4.53 & 4.90 & 4.29 & 4.81 \\
MusicInfuser         & 6.24 & 6.22 & \textbf{6.13} & 5.52 & 3.18 & 3.07 & 2.91 & 3.02 & 4.53 \\
Wan-Dancer           & 5.90 & 5.87 & 5.60 & 5.10 & 4.98 & 5.22 & 5.14 & 4.82 & 5.33 \\
X-Dancer$^{\dagger}$ & 6.17 & 6.01 & 5.73 & 5.05 & 5.89 & 6.50 & 6.64 & 6.16 & 6.02 \\
Ours & \textbf{6.39} & \textbf{6.28} & 5.98 & \textbf{5.72} & \textbf{6.12}
     & \textbf{6.63} & \textbf{6.80} & \textbf{6.40} & \textbf{6.29} \\
\hline
\end{tabular*}
\end{table*}

\begin{table}[t]
\centering
\caption{User-study preference (\%) for EverybodyDance by reference type.
As in Tables~\ref*{tab:main_comparison} and~\ref*{tab:control_ablation}, each value is the percentage of user-study judgments favoring EverybodyDance;
each comparison has 799 judgments for realistic humans and 1,222 for stylized
characters.}
\label{tab:pref_by_type}
\footnotesize
\begin{tabular*}{\linewidth}{@{\extracolsep{\fill}}lcc@{}}
\hline
Compared with & Realistic humans & Stylized characters \\
\hline
\multicolumn{3}{@{}l}{\emph{Baselines (Table~\ref*{tab:main_comparison})}} \\
X-Dancer              & 99.12 & 99.10 \\
MusicInfuser          & 98.12 & 94.11 \\
Wan-Dancer            & 95.49 & 98.77 \\
X-Dancer$^{\dagger}$  & 98.62 & 96.32 \\
\hline
\multicolumn{3}{@{}l}{\emph{Variants (Table~\ref*{tab:control_ablation})}} \\
Raw Coordinates       & 97.75 & 96.24 \\
Raw Coordinates + VAE & 90.99 & 89.53 \\
T2I RoPE              & 87.61 & 86.09 \\
\hline
\end{tabular*}
\end{table}

\subsection{User-Study Protocol}
\label{app:user_study}

Each trial used a three-column interface. The first column showed the
reference image, and the second and third columns showed two dance videos
labeled A and B. Raters received the following instruction:

\begin{quote}
Based on the reference image and accompanying music, select the video with
the better overall result between A and B. Please mainly consider:

\begin{enumerate}
    \item \textbf{Music alignment:} whether the dance responds appropriately
    to the rhythm and section changes of the music, and whether the movement
    intensity and emotion suit the music.
    \item \textbf{Appearance consistency:} whether the character appearance,
    clothing, and visual style are consistent with the reference image.
    \item \textbf{Motion and visual quality:} whether the motion is natural
    and fluent, whether the video is stable, and whether it contains visible
    problems such as body deformation, flickering, or interpenetration.
\end{enumerate}

Considering these aspects together, select either ``A is better'' or
``B is better.''
\end{quote}

For comparisons against MusicInfuser, which does not take a reference image,
the first column instead displayed an instruction card stating: ``No
reference image is provided; appearance consistency does not need to be
considered.''

Method names were hidden, and the assignment of the two methods to A and B
was randomized. Ties were not allowed, and raters could replay the videos.
EverybodyDance was compared separately against four baselines (X-Dancer,
MusicInfuser, Wan-Dancer, and X-Dancer$^{\dagger}$) and three ablation
variants (Raw Coordinates, Raw Coordinates + VAE, and T2I RoPE). All 43
character--music pairs were included in each comparison. Each of the 47
raters therefore completed $43\times7=301$ pairwise trials. Each
EverybodyDance-versus-comparator comparison received $47\times43=2{,}021$
votes, for $14{,}147$ judgments in total.

\section{Additional Inference Details}
\label{app:training_inference}

\subsection{Long-Track Keypoint Generation}
\label{app:long_track}

For music longer than one segment, we generate keypoint sequences segment by segment.
The first segment uses keypoints from the input reference image. For each
subsequent segment $s$, we take the final frame of the preceding generated
keypoint sequence as the new reference,
$x_{\mathrm{ref}}^{(s)}=\hat X^{(s-1)}_{T}$.
We repeat this frame for $T_{\mathrm{ref}}$ frames and encode it using
Equation~\ref*{eq:reference}. The resulting clean reference latent conditions
generation for the corresponding music segment, using the segment sampling
procedure in Section~\ref*{sec:flow}. The generated keypoint sequences are
assembled in temporal order. Only the first-stage reference keypoints are updated;
the original character image remains the reference for video synthesis
(Appendix~\ref{sec:rendering}).

\subsection{Long-Video Generation with Dynamic Overlapping Windows}
\label{sec:rendering}

We use the pretrained pose-conditioned video generation model SCAIL
\citep{yan2026scail} to synthesize the character video from the reference image
$I_{\mathrm{ref}}$ and a pose-conditioning video constructed from the generated
2D keypoint sequence $\hat X$.

\paragraph{Pseudo-3D cylindrical pose rendering for SCAIL.}
To construct the SCAIL pose input, we convert the $T$-frame sequence of
normalized 2D keypoints into a pseudo-3D cylindrical skeleton video through
heuristic depth lifting and signed distance field (SDF) rendering. Given a
canvas of width $W$ and height $H$, we place a pinhole camera at the origin
with a field of view $\theta_{\mathrm{fov}}=55^\circ$ along the longer side,
focal length
$f_{\mathrm{cam}}=\max(H,W)/(2\tan(\theta_{\mathrm{fov}}/2))$, and principal
point $(c_x,c_y)=(W/2,H/2)$. At frame $t\in\{1,\ldots,T\}$, normalized
coordinates $(x_{j,t},y_{j,t})\in[0,1]^2$ map to pixel coordinates
$(\tilde{x}_{j,t},\tilde{y}_{j,t})=(x_{j,t}W,y_{j,t}H)$.

We map our body landmarks to the 14 OpenPose core body joints
\citep{cao2021openpose}, relabel them as $j\in\{1,\ldots,14\}$ from the nose
to the two ankles, and let $\mathcal J_t\subseteq\{1,\ldots,14\}$ denote the
visible joints at frame $t$. Their vertical pixel span is
$\Delta\tilde{y}_t=\max_{j\in\mathcal J_t}\tilde{y}_{j,t}
-\min_{j\in\mathcal J_t}\tilde{y}_{j,t}$. Assuming a canonical 3D body height
$H_{\mathrm{3D}}=1100$, we estimate the scale-adaptive frame depth as
\begin{equation}
 z_t^{\mathrm{raw}}=
 \min\!\left(6000,\max\!\left(1500,
 \frac{f_{\mathrm{cam}}H_{\mathrm{3D}}}{\Delta\tilde{y}_t}\right)\right).
 \label{eq:pseudo3d_raw_depth}
\end{equation}
For degenerate frames with $|\mathcal J_t|<2$ or $\Delta\tilde{y}_t<20$, we
instead set $z_t^{\mathrm{raw}}=3.5f_{\mathrm{cam}}$. We smooth the frame depth
over $\mathcal N_t=\{\tau\in\{1,\ldots,T\}:|\tau-t|\leq2\}$ and back-project
each visible joint as
\begin{equation}
\begin{aligned}
 \bar z_t
 &=\frac{1}{|\mathcal N_t|}
   \sum_{\tau\in\mathcal N_t}z_\tau^{\mathrm{raw}},\\
 \mathbf X_{j,t}
 &=\left(
   \frac{\tilde{x}_{j,t}-c_x}{f_{\mathrm{cam}}}z_{j,t},
   \frac{\tilde{y}_{j,t}-c_y}{f_{\mathrm{cam}}}z_{j,t},
   z_{j,t}\right)^\top,
\end{aligned}
\label{eq:pseudo3d_backprojection}
\end{equation}
where $z_{j,t}=\bar z_t+\Delta z_j$. The predefined anatomical offset
$\Delta z_j\in[-30,10]$, measured relative to the neck, induces a
front-to-back self-occlusion order.

Let $\mathcal L_t$ contain the valid limbs among the 13 OpenPose
body connections, forming the torso, arm chains from the neck to both wrists,
leg chains from the neck to both ankles, and the neck-to-head connection.
We model each limb as a capsule-shaped cylindrical primitive of radius
$r=21.5$. Its scene SDF is
\begin{equation}
 d_t(\mathbf q)=
 \min_{\substack{(i,j)\in\mathcal L_t\\\lambda\in[0,1]}}
 \left\|\mathbf q-(1-\lambda)\mathbf X_{i,t}
              -\lambda\mathbf X_{j,t}\right\|_2-r,
 \label{eq:pseudo3d_sdf}
\end{equation}
where $\lambda$ parameterizes the line segment between the two joints. We
render this SDF using perspective sphere tracing, SCAIL's limb palette, and
depth-attenuated Blinn--Phong shading. Finally, we overlay each valid
21-joint 2D DWPose hand skeleton \citep{yang2023dwpose} after jump-outlier
filtering.

For all SCAIL-based results, we combine SCAIL
with the acceleration LoRA provided by LightX2V \citep{lightx2v} and use six
denoising steps. The same configuration is used for EverybodyDance,
X-Dancer$^{\dagger}$, and all keypoint-sequence generation variants in
Section~\ref*{sec:control_ablation}.
We extend video generation using noise rescheduling and dynamic overlapping
windows, without additional training. Local noise is shuffled and reused
following FreeNoise \citep{qiu2024freenoise} to promote long-range appearance
consistency. At each denoising step, we aggregate predictions from overlapping
windows using linear blending weights and normalize by the sum of weights
before updating the global video latent sequence. Window boundaries are shifted
across steps to improve cross-window consistency and reduce boundary artifacts,
similar to Sonic's window-shifting strategy \citep{ji2025sonic}.
The context length and overlap are $81$ and $48$ video frames, corresponding
to $21$ and $12$ video-latent frames, respectively.

The same music-conditioned generator, trained on real-human dance keypoints,
is reused across character types without anime- or animal-specific
music--dance pairs. Cross-style video synthesis is provided by pretrained SCAIL;
the music-conditioned generator and trajectory-image VAE are not retrained
for different character types.

\subsection{Template Reference Keypoints}
\label{app:template_reference}

Keypoint detection can fail on some stylized reference images. In this case,
we use a fixed template skeleton of a person in a natural standing pose as the
reference keypoints $x_{\mathrm{ref}}$, and keep the trained model and the
sampling procedure of Section~\ref*{sec:flow} unchanged. The generated keypoint sequence starts from this canonical standing pose, and
SCAIL synthesizes the target character directly from the reference image
without requiring initial pose alignment (Video~S2). An example of the template
skeleton is shown in the first frame of the pose-conditioning video in the
middle column of Video~S2.

\section{Metrics for VAE Reconstruction}
\label{app:vae_metrics}

This section defines the reconstruction metrics reported in
Table~\ref*{tab:codec_ablation}. Let $\mathbf p_{n,k,t}$ and
$\hat{\mathbf p}_{n,k,t}$ denote the target and reconstructed positions of
landmark $k$ at frame $t$ in sequence $n$. Before measuring coordinate error,
we invert any coordinate transformations and express both positions in the
same normalized full-image coordinate system.

\subsection{Global and Anchor-Aligned L2}
\label{app:aligned_l2}

\paragraph{Overall L2.}
Let $\mathcal V_{\mathrm{all}}$ be the landmark--frame instances retained
by the coordinate-error evaluation mask. Overall L2 is the mean Euclidean
distance between reconstructed and target positions:
\begin{equation}
    \mathrm{L2}_{\mathrm{overall}}
    =\frac{1}{|\mathcal V_{\mathrm{all}}|}
      \sum_{(n,k,t)\in\mathcal V_{\mathrm{all}}}
      \left\|\hat{\mathbf p}_{n,k,t}-\mathbf p_{n,k,t}\right\|_2.
    \label{eq:metric_overall_l2}
\end{equation}
This is a mean of Euclidean distances, not a mean squared error.
Body L2 applies the same average to the 20 body landmarks.

\paragraph{Hand and face aligned L2.}
For each landmark, let $a(k)$ denote its anchor: the same-side wrist for
hand landmarks and the nose tip for facial landmarks. We subtract the
predicted anchor from the predicted landmark and the target anchor from
the target landmark:
\begin{equation}
    \begin{aligned}
        \hat{\boldsymbol\delta}_{n,k,t}
          &=\hat{\mathbf p}_{n,k,t}
            -\hat{\mathbf p}_{n,a(k),t},\\
        \boldsymbol\delta_{n,k,t}
          &=\mathbf p_{n,k,t}-\mathbf p_{n,a(k),t}.
    \end{aligned}
    \label{eq:metric_anchor_relative}
\end{equation}
For $P\in\{\mathrm{hand},\mathrm{face}\}$, the aligned error is
\begin{equation}
    \mathrm{L2}_{P,\mathrm{aligned}}
    =\frac{1}{|\mathcal V_P|}
      \sum_{(n,k,t)\in\mathcal V_P}
      \left\|\hat{\boldsymbol\delta}_{n,k,t}
                    -\boldsymbol\delta_{n,k,t}\right\|_2,
    \label{eq:metric_aligned_l2}
\end{equation}
where $\mathcal V_P$ denotes the retained instances for that part.
This removes anchor-position error and measures local relative geometry.
All L2 scores are reported in units of $10^{-3}$ normalized image
coordinates; the table therefore lists $10^3$ times the errors above.

\subsection{Missing-Landmark Detection}
\label{app:missing_metrics}

Missing or invalid landmarks form the positive class. Let $v^a_{n,k,t}$
and $\hat v^a_{n,k,t}$ denote target and predicted validity for coordinate
axis $a$. A landmark is missing if either axis is missing:
\begin{equation}
    \begin{aligned}
        m_{n,k,t}&=1-v^x_{n,k,t}v^y_{n,k,t},\\
        \hat m_{n,k,t}&=1-\hat v^x_{n,k,t}\hat v^y_{n,k,t}.
    \end{aligned}
    \label{eq:metric_landmark_missing}
\end{equation}
Thus, detection statistics are computed at landmark level, although the
model predicts axis-level validity.
For configurations with a null output, missingness is determined from that
output. For Global MSE and Global CE, confidence thresholds are selected
to maximize Missing F1 on separate validation sequences and then kept fixed
during evaluation.

Let $\mathcal E$ be the common set of landmark--frame instances retained for
missingness evaluation. Over this set, TP counts truly missing landmarks
predicted as missing; FP counts truly valid landmarks predicted as missing;
and FN counts truly missing landmarks predicted as valid. Formally,
\begin{equation}
    \begin{aligned}
        \mathrm{TP}&=\sum_{\mathcal E}m\,\hat m,\\
        \mathrm{FP}&=\sum_{\mathcal E}(1-m)\,\hat m,\\
        \mathrm{FN}&=\sum_{\mathcal E}m\,(1-\hat m),
    \end{aligned}
    \label{eq:metric_missing_counts}
\end{equation}
where the indices $(n,k,t)$ are suppressed for readability.
For nonzero denominators, Missing F1 and Missing FNR, reported as
percentages, are
\begin{equation}
    \begin{aligned}
        \mathrm{F1}_{\mathrm{miss}}(\%)
          &=100\frac{2\mathrm{TP}}
            {2\mathrm{TP}+\mathrm{FP}+\mathrm{FN}},\\
        \mathrm{FNR}_{\mathrm{miss}}(\%)
          &=100\frac{\mathrm{FN}}{\mathrm{TP}+\mathrm{FN}}.
    \end{aligned}
    \label{eq:metric_missing_f1_fnr}
\end{equation}
Missing F1 balances precision and recall for the missing class. Missing FNR
measures the fraction of truly missing landmarks incorrectly treated as
valid, and equals $100(1-\mathrm{Recall}_{\mathrm{miss}})$ when recall is
expressed on a $[0,1]$ scale.

\subsection{Evaluation Masks}
\label{app:metric_masks}

Each metric uses one ground-truth mask shared by all four training
configurations. Missingness metrics use the global/local validity-agreement
mask. L2 metrics use the jointly valid landmarks whose ground-truth local
encoding round-trips to image space within $0.01$; the sets
$\mathcal V_{\mathrm{all}}$, $\mathcal V_{\mathrm{hand}}$, and
$\mathcal V_{\mathrm{face}}$ refer to these instances. Both masks depend only on
ground-truth annotations, not on model predictions, so every column of
Table~\ref*{tab:codec_ablation} compares the four configurations on the same
instances. On the 512 holdout sequences, the L2 mask keeps 8,309,075 of the
8,320,068 jointly valid landmark instances (99.87\%): all body and facial
instances and 99.65\% of hand instances. The excluded hand instances exceed the
clipping range of the hand-coordinate map (Appendix~\ref{app:hand_coordinates}).

\section{Implementation of Coordinate-Based Generation Variants}
\label{app:generation_variants}

This section details the coordinate-based comparisons in
Section~\ref*{sec:control_ablation}.

\subsection{Shared DiT Backbone}
\label{app:shared_dit}

Raw Coordinates, Raw Coordinates + VAE, and EverybodyDance use the same
NextDiT backbone as in Section~\ref*{sec:exp_setup}, with 26 Transformer
blocks, hidden dimension 2304, and patch size 1.
To keep the comparison as controlled as possible, all Transformer blocks, the shared-time RoPE strategy, music conditioning,
training objective (the flow-matching loss in Equation~\ref*{eq:flow_loss}),
optimizer, and flow-matching schedule are the same; only the input and output projections are adapted to each representation,
with 72, 128, and 166 channels for EverybodyDance, Raw Coordinates + VAE, and
Raw Coordinates, respectively.

\subsection{Continuous 1D Coordinate VAE}
\label{app:raw_coordinate_vae}

The coordinate VAE adapts the Bailando-based temporal-convolutional
encoder--decoder described in Section~\ref*{sec:control_ablation}.
Its encoder has three stride-2 temporal downsampling layers, matching the
temporal stride of our trajectory-image VAE, and dilated convolutional residual
blocks with dilation rates 1, 3, and 9.
The model contains approximately 43.6M parameters, matching the approximate
scale of our trajectory-image VAE.

We remove the discrete VQ codebook and replace it with a diagonal Gaussian
posterior. For a coordinate sequence $X$, the encoder predicts the posterior mean
$\mu_\phi(X)$ and log-variance $\log\sigma_\phi^2(X)$.
During VAE training, continuous latents are sampled using reparameterization:
\begin{equation}
    \begin{aligned}
        \epsilon&\sim\mathcal N(0,I),\\
        Z_{\mathrm{1D}}
          &=\mu_\phi(X)+\sigma_\phi(X)\odot\epsilon.
    \end{aligned}
    \label{eq:raw_vae_reparameterization}
\end{equation}
Here $\odot$ denotes elementwise multiplication. Training includes KL
regularization of the posterior. The resulting continuous latent
representation interfaces directly with the flow-matching generator,
without a codebook lookup or vector quantization.

\subsection{Reference Conditioning}
\label{app:variant_grids}

As in EverybodyDance, each variant conditions on the same reference pose,
prepended to its main sequence along the temporal axis. The reference stays
clean and serves only as conditioning; flow interpolation and prediction losses
apply only to the main sequence.

\end{document}